\documentclass{article}
\usepackage{spconf,amsmath,graphicx}
\usepackage{booktabs}
\usepackage{tikz}
\usepackage{pgfplots}
\usepackage{amssymb}
\usepackage{subfigure}
\usepackage{mathtools}
\usepackage{float}

\DeclarePairedDelimiterX{\norm}[1]{\lVert}{\rVert}{#1}

\title{Contrastive predictive coding for Anomaly Detection in Multi-variate Time Series Data}
\name{Author(s) Name(s)\thanks{Thanks to XYZ agency for funding.}}
\address{Author Affiliation(s)}
\twoauthors
 {Theivendiram Pranavan, Terence Sim}
	{School of Computing, NUS\\
	Singapore}
 {Arulmurugan Ambikapathi, Savitha Ramasamy}
	{Institute for Infocomm Research, (I\textsuperscript{2}R), A*STAR\\
	Singapore}

\begin{document}
%
\maketitle
\begin{abstract}
    Anomaly detection in multi-variate time series (MVTS) data is a huge challenge as it requires simultaneous representation of long term temporal dependencies and correlations across multiple variables. More often, this is solved by breaking the complexity through modeling one dependency at a time. In this paper, we propose a Time-series Representational Learning through Contrastive Predictive Coding (TRL-CPC) towards anomaly detection in MVTS data. First, we jointly optimize an encoder, an auto-regressor and a non-linear transformation function to effectively learn the representations of the MVTS data sets, for predicting future trends. It must be noted that the context vectors are representative of the observation window in the MTVS. Next, the latent representations for the succeeding instants obtained through non-linear transformations of these context vectors, are contrasted with the latent representations of the encoder for the multi-variables such that the density for the positive pair is maximized. Thus, the TRL-CPC helps to model the temporal dependencies and the correlations of the parameters for a healthy signal pattern. Finally, fitting the latent representations are fit into a Gaussian scoring function to detect anomalies. Evaluation of the proposed TRL-CPC on three MVTS data sets against SOTA anomaly detection methods shows the superiority of TRL-CPC.

\end{abstract}
\begin{keywords}
Contrastive Predictive Coding(CPC), Univariate and Multi-variate Time Series (MVTS), Representation Learning, Slow Features
\end{keywords}
\section{Introduction}
\label{sec:intro}
With the recent widespread real-time adoption of IoT in several applications, there is an abundance of MVTS data. This also opens up opportunities to pre-empt failures in order to enable timely intervention, thereby, avoiding catastrophic events. As the failure events in real-world environments are rare, there is a need to detect anomalies from unlabelled data. Anomaly detection is defined as detecting data segments that deviate from the underlying distribution of the data. Detecting anomalies in a MVTS data set is challenging due to the need to model both the temporal correlations, and the correlations across the multi-variate channels\cite{teng2010anomaly, malhotra2015long}. There has been consistent research towards detecting anomaly from time-series data \cite{braei2020anomaly}. With the recent developments in deep learning and their successes on image data sets, there is increased attention to develop deep learning methods for anomaly detection from MVTS \cite{audibert2020usad,li2019mad}. Despite these progress, their anomaly detection accuracies are not as good as multi-channel univariate autoencoders \cite{garg2021evaluation}.
\begin{figure}
    \centering
    \includegraphics[width=0.5\textwidth]{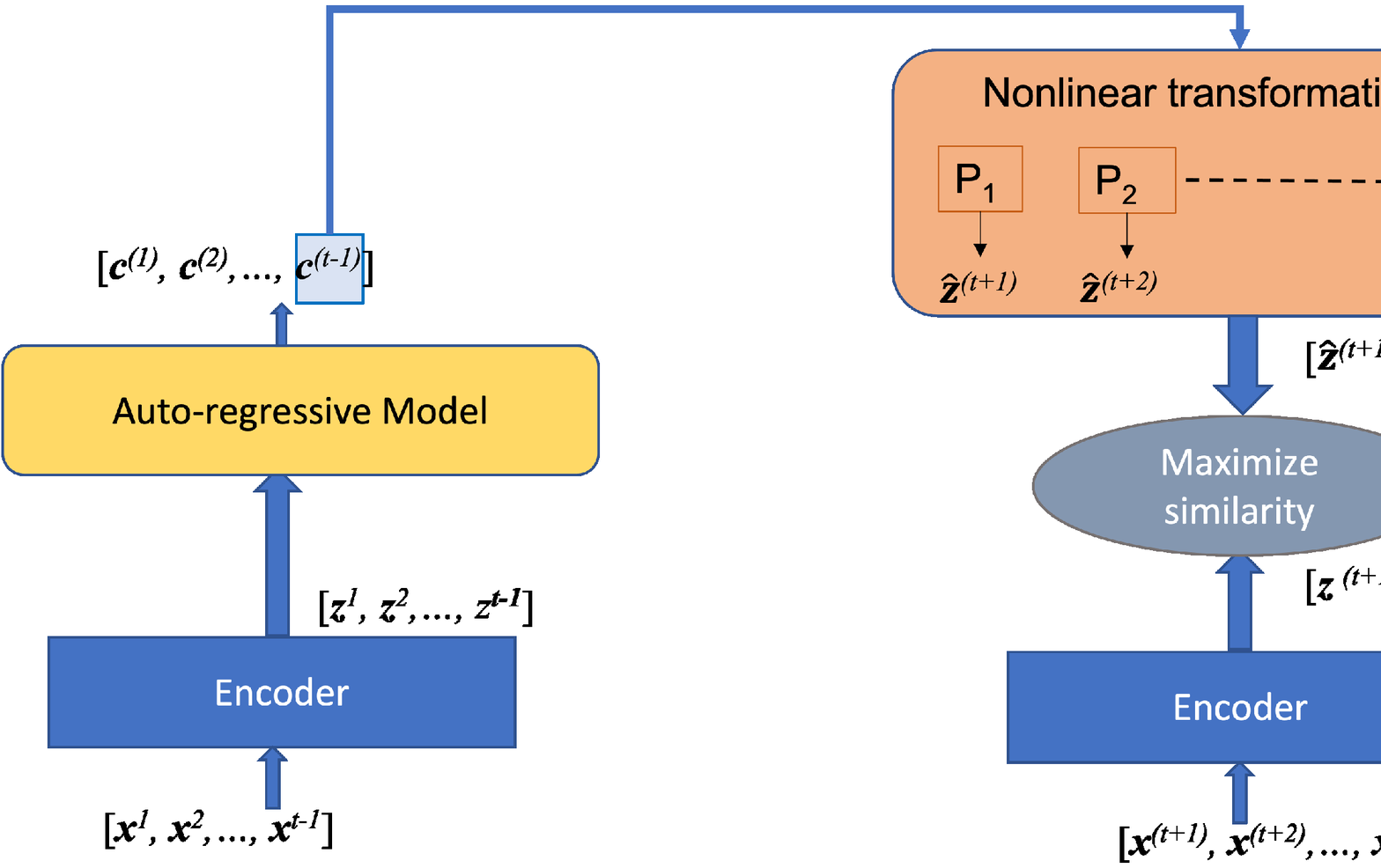}
    \caption{ The figure shows the case where $t-1$ time steps predicts $k$ future time steps.  $\mathbf{x}^{1}\ldots  \mathbf{x}^{t-1}, \mathbf{x}^{t+1} \ldots \mathbf{x}^{t+k}$ are inputs; $E$ - encoder; $\mathbf{z}^{1} \ldots \mathbf{z}^{t-1}, \mathbf{z}^{t+1} \ldots \mathbf{z}^{t+k}$ - latent representations of $\mathbf{x}^{1}\ldots  \mathbf{x}^{t-1}, \mathbf{x}^{t+1} \ldots \mathbf{x}^{t+k}$; $AR$ - autoregressor; $\mathbf{c}^{1} \ldots \mathbf{c}^{t-1}$ - contexts(For example; $\mathbf{c}^i$ accumulates history of $\mathbf{z}^{1} \ldots \mathbf{z} ^ {i}$); $P_{1 \ldots k}$ - prediction networks. $AR$ and $E$ are shared across the network. The whole model is trained end to end using a toy task which is a binary classification. It is noted that $\mathbf{c}^{t-1}$ is used for predictions not the previous contexts.}
    \label{fig:cpc}
\end{figure}

\begin{table*}
\centering
\caption{CPC and Contrastive Learning previous work summary}
\label{tab_summary}
\begin{tabular}{|p{5cm}|p{7cm}|p{2cm}|}
\toprule
Area &  Datasets & Previous Work\\
\midrule
Computer Vision(Images and Videos) & Imagenet, CIFAR10, CIFAR100, Caltech-101,VOC2007 & \cite{oord2018representation,henaff2020data,tian2019contrastive,chen2020simple}\\
Reinforcement Learning & DeepMind Lab, DMControl100k, atari100k & ~\cite{laskin2020curl,oord2018representation}\\
Natural Language Processing(NLP) & BookCorpus, SentEval & ~\cite{oord2018representation,giorgi2020declutr}\\
Univariate Time Series(Audio) & LibriSpeech~\cite{panayotov2015librispeech}& ~\cite{oord2018representation}\\
Multivariate Time Series & Damadics, WADI, SKAB & - \\
\bottomrule
\end{tabular}
\label{tab-01}
\end{table*}

Recently, contrastive predictive coding \cite{oord2018representation, henaff2020data} approaches offer better representational abilities than traditional deep learning models and are hence, good candidates to learn from MVTS \cite{eldele2021time}. However, their ability to represent temporal and channel correlations for anomaly detection is not fully known. In this paper, we propose a Time-series Representational Learning through Contrastive Predictive Coding (TRL-CPC) model for anomaly detection from MVTS \ref{fig:cpc}. The TRL-CPC comprises of an encoder, an auto-regressor, and a non-linear transformation function. First, the encoder and the auto-regressor learns the representation of the sequence of the time-series in the observation window to generate the context vectors. The nonlinear transformation function, which can be approximated through a single hidden layer perceptron network, then transforms these context vectors to estimate the latent vectors for the sequence of time-series in the prediction window. This estimated latent vectors is then contrasted with the actual latent vectors from the encoder for better representation of the relationship between the observation and prediction window. All the three components, viz., the encoder, the auto-regressor and the non-linear transformation function are jointly optimized such that the density of the positive pair (time-series sequence of observation Vs prediction window) is maximized. We demonstrate the anomaly detection capabilities of the proposed TRL-CPC on 2 publicly available MVTS data set, namely, Skoltech Anomaly benchmark \cite{skab} and WADI \cite{ahmed2017wadi} with $36.02$\% and $5.79$\% of anomalies, respectively. These data sets are chosen with extremities in the anomaly \%. Finally, we apply the proposed method on a real-world data set. Performance studies show the superior anomaly detection capabilities of the proposed TRL-CPC.


\section{Related Work}
In this section, we briefly review the literature on Constrastive Predictive Coding. Predictive Coding\cite{elias1955predictive,huang2011predictive} is an important brain function which is used to compare the actual sensory input against the predicted sensory input. The errors in the prediction are used to update and revise the internal model of the brain. The discovery of Predictive Coding rooted from unconscious inference~\cite{von1867treatise}. Initial ideas focus on filling in visual information, but the idea is universal to other sensory inputs as well. Recent fMRI study shows that predictability of a stimuli influences the responses of primary visual cortex~\cite{alink2010stimulus}. Several previous works show the existence of predictive coding through brain imaging studies\cite{alink2010stimulus,den2010striatal,smith2010nonstimulated,todorovic2012repetition}.

Despite the early discovery of predictive coding in neuroscience, the idea of predictive coding is at the inception level in machine learning until recently. Contrastive Predictive Coding~\cite{oord2018representation,henaff2020data} is a self-supervised representational learning technique inspired by Predictive Coding. In a CPC, the loss is applied in the representational space instead of the output space. To this end, the main objective of a CPC model is to distinguish represenations, instead of optimizing output loss function. CPC has produced impressive results due to its ability to find generalized representations suitable for various downstream tasks.

With the recent development in contrastive approaches that have huge potential in representative learning, they are good candidates for representing multivariate time series data sets \cite{eldele2021time}. However, their potential for anomaly detection is less explored. In the next section, we propose a multi-variate  Time-series Representational Learning through Contrastive Predictive Coding (TRL-CPC), towards anomaly detection. Table \ref{tab_summary} summarizes some recent literature in CPC.

\section{CPC Framework for Anomaly Detection in Time Series}

In this section, we introduce the problem of anomaly detection from multi-variate time series data, and then present the proposed TRL-CPC.
\subsection{Anomaly Detection Problem Formulation}
Let $X=\{\mathbf{x}^1, \ldots, \mathbf{x}^t, \ldots, \mathbf{x}^T\} \subset \mathbb{R}^{m}$ be a {\color{black}discrete time-series data set} with $T$ time instances, and $\mathbf{x}^t = \left[x_1^t, \ldots, x_m^t\right]^T$ $\in \mathbb{R}^{m}$ is the {\color{black}m-dimensional} multi-variate signal at time instance $t$. The objective of anomaly detection is to learn the representations of this multi-variate time-series (MVTS) data set, such that any multi-variate sample that has a distinct representation in this time-series data set can be identified as an anomaly. This can be solved through modeling the representation of the time-series samples $D = \{\mathbf{x}^1, \ldots, \mathbf{x}^{t-1}\}$ in the observation window to predict the representations for another set of time-points (prediction window), defined as $S = \{\mathbf{x}^{t+k}, \ldots, \mathbf{x}^{t+l}\}$, where $k \geq 0, ~l\geq k$. In this section, we propose a representation learning approach through contrastive predictive coding towards anomaly detection.

\subsection{CPC Framework for Time Series} \label{cpc-framework}
The TRL-CPC comprises of an encoder, an auto-regressive model and a non-linear transformation model, as shown in Fig. \ref{fig:cpc}. Given the time-series in the observation window $D=\{\mathbf{x}^1, \ldots, \mathbf{x}^{t-1}\}$, the encoder and the autoregressive model learns the representation of the multi-variate time series for the observation window through the latent representations $\{\mathbf{z}^1, \ldots, \mathbf{z}^{t-1}\} \subset \mathbb{R}^{n}, ~n<m,$ to generate a context vector $\{\mathbf{c}^{1}, \ldots, \mathbf{c}^{t-1}\} \subset \mathbb{R}^{a}, ~a>0$, for the given prediction window. The context vectors for the prediction window is then non-linearly transformed to reconstruct the latent vectors of the prediction window, i.e. $\{\widehat{\mathbf{z}}^{t+1}, \ldots, \widehat{\mathbf{z}}^{t+k}$. Simultaneously, the latent representations of the prediction window $\{\mathbf{z}^{t+1}, \ldots, \mathbf{z}^{t+k} \}$  is obtained through the encoder. Finally, the similarity between the reconstructed latent features of the prediction window and the actual latent features are maximized towards efficient representation of the temporal and parameter correlations for positive pair (the prediction window that immediately follows the previous window). Thus, the encoder and the auto-regressive model are jointly optimized for representational learning of the MVTS towards anomaly detection.

It must be noted that the Encoder ($E$) and autoregressive model ($AR$) are shared across the whole network. The latent representation of the observation window is given by
\begin{eqnarray}
    \{\mathbf{z}^1, \ldots, \mathbf{z}^{t-1}\} = E(\mathbf{x}^1, \ldots, \mathbf{x}^{t-1}).\label{encoder-eq}
\end{eqnarray}
Thereafter, the context vectors of the prediction window are obtained through the autoregressive model ($AR$) as:
\begin{eqnarray}
\{\mathbf{c}^1, \ldots, \mathbf{c}^{t-1} \}= AR\left(\{\mathbf{z}^1, \ldots, \mathbf{z}^{t-1}\}\right).
\end{eqnarray}

The latent representations of the prediction window are reconstructed through a nonlinear transformation of these context vectors as:
\begin{eqnarray}
 \{\widehat{\mathbf{z}}^{t+1}, \ldots, \widehat{\mathbf{z}}^{t+i}\} = P_i(\mathbf{c}^{t-1}); i = t+1,\ldots,t+k
\end{eqnarray}
where $P_k$ is a nonlinear transformation for each multi-variate sample in the prediction window that can be accomplished through a neural network such as a single hidden layer multi-layer perceptron.

Simultaneously, the actual latent representations for the samples in the prediction window can be obtained through the encoder as:
\begin{eqnarray}
  \{{\mathbf{z}}^{t+1}, \ldots, {\mathbf{z}}^{t+k}\} = E(\mathbf{x}^{t+1}, \ldots, \mathbf{x}^{t+k})\label{encoder-eq1}
\end{eqnarray}

The encoder, auto-regressive model and the nonlinear transformation function are now jointly optimized to maximize similarity between the reconstructed and actual latent vectors of the samples in the prediction window. In order to learn this, the positive pair (prediction window subsequent to the observation window) and negative pairs (any other prediction windows away from the observation window) are chosen from the batch of time-series samples. The objective of the optimization is to maximize the density of the positive pair. Hence, we compute the scalar score for the prediction window using log-bilinear model, as shown below:
\begin{equation}
    f_i(\mathbf{x}_{t+i}, \mathbf{c}^{t-1}) = \exp(\mathbf{z}_{t+i}^T P_i(\mathbf{c}^{t-1}));~i=1,\ldots,k
    \label{eq:fu_p}
\end{equation}

These predictions are then used to optimize the density ratio of the positive pair with a loss based on InfoNCE. Given a set of $M$ examples (batch size) with $M-1$ negative pairs and one positive pair, the following loss is jointly optimized:
\begin{eqnarray}
    \mathbb{L}_M &=& - \mathbb{E}_G \left[\log \dfrac{f_i(\mathbf{x}_{t+i}, \mathbf{c}^{t-1})}{\Sigma_{\mathbf{x}_j \in X} f_i(\mathbf{x}_j, \mathbf{c}^{t-1})}\right]\\\nonumber i&=&1,\ldots,k; ~j=1,\ldots,M
\end{eqnarray}

Having a larger $M$ helps to improve performance. In our experiments, we use $M=8, 16, 64$, with one positive pair per batch.

\subsubsection{Anomaly Detection using Gaussian}
\label{m1}

Sets $R_1$ and $R_2$ contains representations (as defined in Eqn \ref{eq-2}) of the training set $V$ and the testing set $W$, respectively. That is,

\begin{equation}
    \begin{aligned}
         R_1 &= \{\mathbf{z} \: | \: \forall \mathbf{x} \in V, \mathbf{z} = E(\mathbf{x}) \} \\
         R_2 &= \{\mathbf{z} \: | \: \forall \mathbf{x} \in W, \mathbf{z} = E(\mathbf{x}) \}
    \end{aligned}
    \label{eq-2}
\end{equation}

Set $R_1$ is fitted to a Gaussian distribution ($G$) with mean $\boldsymbol{\mu} \in \mathbb{R}^{n}$, and covariance $\boldsymbol{\Sigma} \in \mathbb{R}^{n \times n}$. Then, for the set $R_2$, the likelihood to the distribution $G$ is calculated using probability density function $h(\cdot)$. Then each of these probabilities are used as thresholds to determine whether a sample is an anomaly (label $1$) or not (label $0$). For every probability ($p$), anomaly is decided using Eqn \ref{eq-3}. For each probability $p$, corresponding labels for test set is calculated and the $F1$ score is recorded. Among all $F1$ scores, the best $F1$ is selected.

\begin{equation}
   \text{label}(\mathbf{x}) = [h(E(\mathbf{x})) \leq p] =
  \begin{cases}
   1 \\
    0\\
  \end{cases}
  \label{eq-3}
\end{equation}

\begin{table*}
\centering
  \caption{Damadics, WADI, and SMD details. Here the number of samples corresponds to the number of timestamps. Training samples do not contain any anomalies. Anomalies are only present in the testing data. \\}
  \label{tab:datasets}
  \begin{tabular}{|c|c|c|c|c|}
    \toprule
    Dataset & Number of Training Samples & Number of Testing Samples & Anomalies(\%) & Channels\\
    \midrule
    Damadics & $518400$ & $259200$ & $16.96$ & $32$\\
    WADI & $1209601$ & $172801$ &  $5.79$ & $123$\\
    SKAB \cite{skab} & $9401$ &  $34282$&  $36.02$ & $8$\\
    \bottomrule
  \end{tabular}
\end{table*}

\section{Experiments and Results}
In this section, we evaluate the anomaly detection performance of the proposed TRL-CPC using the data sets listed in Table \ref{tab:datasets}. It can be observed from the table that the WADI and SKAB have several channels with the least \% of anomalies and the least number of channels and highest \% of anomalies, respectively. The standard $F1$ scores calculated using Eq. \ref{metrics} ($TP$: True Postives, $FP$: False positives and $FN$: false negatives), are used in evaluation.
\begin{equation}
    F1 = \dfrac{2*Pr*Re}{Pr+Re};~Pr = \dfrac{TP}{TP+FP}, Re = \dfrac{TP}{TP+FN}, 
    \label{metrics}
\end{equation}

\begin{figure}
    \centering
    \includegraphics[scale=0.3]{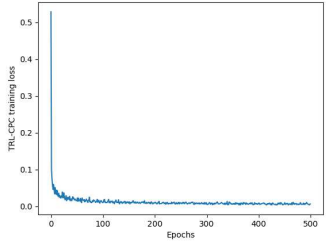}
    \caption{Damadics dataset - TRL-CPC join optimization loss}
    \label{fig:training_loss}
\end{figure}

In our work, the observation window and the prediction window are 10 timesteps each. Here, each timestep can either represent a particular timestamp or a collection of timestamps. The number of layers in the encoder is directly proportional to the number of channel, however, we use a 2 layer $AR$ model in all our experiments. A learning rate of $0.001$ and Adam optimizer is used to train the TRL-CPC model. Fig \ref{fig:training_loss} shows TL-CPC model convergence in the training.


\begin{figure}

\centering
\begin{subfigure}
\centering
\includegraphics[width=.35\textwidth]{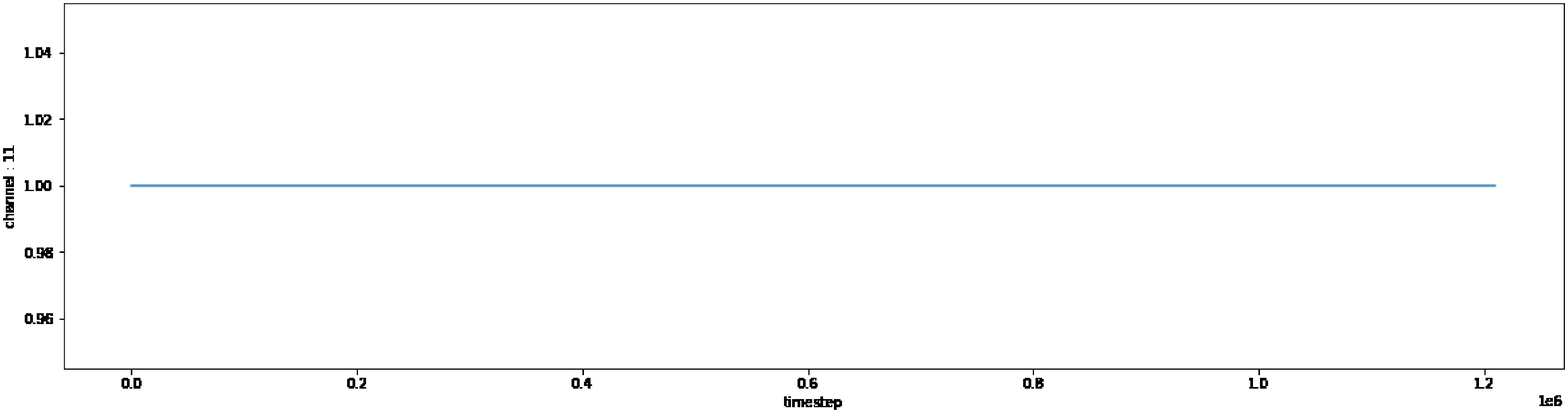}
\end{subfigure}
\vfill
\begin{subfigure}
\centering
\includegraphics[width=.35\textwidth]{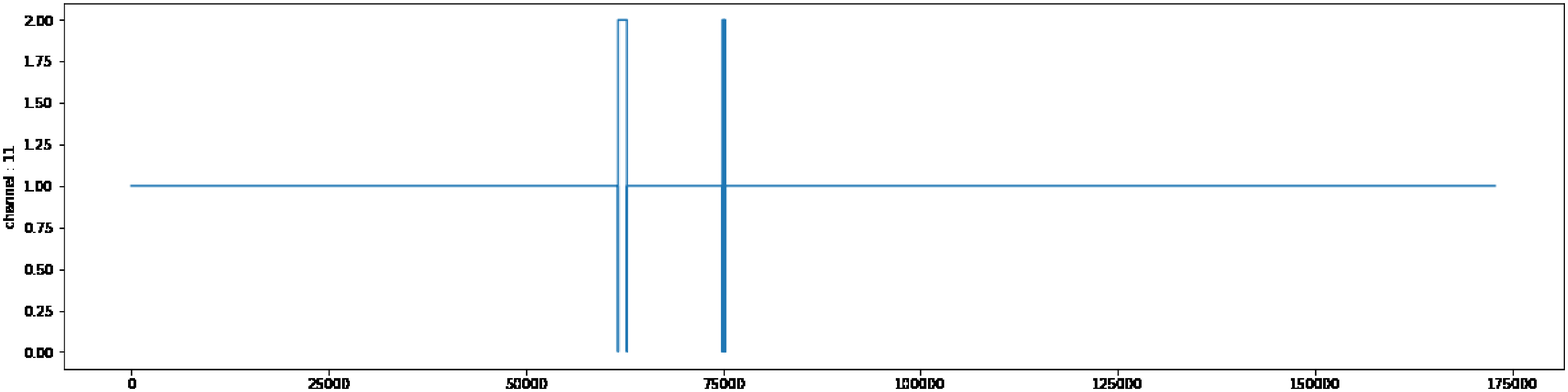}
\end{subfigure}

\caption{The first figure shows the training data which contains a channel which is constant in the training data. But the same channel in testing data(the second figure) has anomalies. Proposed TRL-CPC has some limitation in detecting these anomalies.}
\label{fig:wadi}
\end{figure}

\subsection{Results}
Table \ref{results} presents the results of our proposed TRL-CPC, in comparison with the State-of-the-art (SOTA) methods. From the table, it can be observed that the proposed TRL-CPC is efficient in detecting anomalies. It must be noted that the closest contender is the Univariate Autoencoder, which has been reported as the winning candidate among a number of other methods, for anomaly detection from MVTS \cite{garg2021evaluation}. In such a model, each channel is modelled through a univariate autoencoder to detect anomalies, and is computationally intensive. The proposed TRL-CPC outperforms this UAE in detecting anomalies on the Damadics and SKAB data sets by 0.07 and 0.16, respectively. 




The main limitation of the TRL-CPC is that it is not able to find anomalies when a channel is constant in the training data, but is fluctuating in the testing data. For example, the channel $11$ of the WADI data set is shown in Fig \ref{fig:wadi} , where the training data set is constant, but the testing data is fluctuating.
\begin{table}
\centering
\begin{tabular}{|c|c|c|c|}

\toprule

Method & Damadics & WADI & SKAB \\
\midrule
Conv-AE & - & - & $0.78$ \\
AE & - & $0.23$ & - \\
LSTM-VAE & $\it 0.60$ & $0.23$ & $\it 0.54$ \\
DAGMM & - & $0.12$& - \\
OmniAnomaly & $\it 0.14$ & $0.23$ & $\it 0.46$ \\
USAD \cite{audibert2020usad} & - & $0.23$ & - \\
UAE * \cite{garg2021evaluation} & $0.53$ & $0.35$ & $0.54$ \\

TRL-CPC (Ours) & $0.60$ & $0.25$ & $0.70$\\
\bottomrule
  
\end{tabular}
\caption{AE - AutoEncoder; UAE-Univariate AutoEncoder. $F_1$ scores are compared against SOTA methods. * refers to the models constructed by treating every feature independently. Italicized scores are composite $F$ scores ($F_{c1}$) which are usually higher than traditional $F_1$ score. }
 \label{results}
\end{table}

\section{Conclusion and Future Works}

We propose a representation learning technique for anomaly detection from MVTS datasets. An autoencoder, a auto-regressor and a fully connected network are jointly optimized to learn the repreesntations of MVTS, towards detecting anomalies. Performance studies show the anomaly detection efficiency of the TRL-CPC, and also highlights the main drawback. Future works include improving the represenational learning for anomaly detection from MVTS through improving the encoder, autoregressor and the nonlinear function approximator.


\vfill\pagebreak

\newpage
\bibliographystyle{IEEEbib.bst}
\bibliography{ms}

\end{document}